\documentclass[a4paper,fleqn]{cas-dc}
\usepackage[numbers]{natbib}
\usepackage{graphicx}
\usepackage{multirow}
\usepackage{subfig}
\usepackage{float}
\usepackage{enumerate}
\usepackage{amsfonts}
\usepackage{amsmath}
\usepackage{algorithm}
\usepackage{algorithmicx}
\usepackage{textcomp}

\usepackage{booktabs}
\usepackage{amsmath,amssymb,amsfonts}
\usepackage[noend]{algpseudocode}
\usepackage{xcolor}
\usepackage{threeparttable}
\usepackage{verbatim}

\emergencystretch=\maxdimen
\newcommand{\dt}[1]{\iffalse{#1}\fi}  
\def\tsc#1{\csdef{#1}{\textsc{\lowercase{#1}}\xspace}}
\tsc{WGM}
\tsc{QE}
\tsc{EP}
\tsc{PMS}
\tsc{BEC}
\tsc{DE}

\begin{document}
\let\WriteBookmarks\relax
\def\floatpagepagefraction{1}
\def\textpagefraction{.001}
\shorttitle{Cross-View-Prediction: Exploring Contrastive Feature for Hyperspectral Image Classification}
\shortauthors{Wu \MakeLowercase{\textit{et al.}}}

\title [mode = title]{Cross-View-Prediction: Exploring Contrastive Feature for Hyperspectral Image Classification}

\author[1]{Anyu Zhang}
\address[1]{Soochow University, School of Rail Transportation, Suzhou, ,China}

\author[2]{Haotian Wu}
\address[2]{Imperial College London, Department of Electrical and Electronic Engineerin, London, UK}

\author[3]{Zeyu Cao}
\address[3]{Zhejiang University, College of Electrical Engineering, Hangzhou, China}

\begin{abstract}
This paper presents a self-supervised feature learning method for hyperspectral image classification. Our approach firstly tries to construct two different views of the raw hyperspectral image through a cross-representation learning method. Specifically, four cross-channel-prediction-based augmentation methods are designed for the view construction. And then, better representative features are learned by maximizing mutual information and minimizing conditional entropy between different views from a contrastive network. Features extracted by this `cross-view-prediction' style get the state-of-the-art performance of unsupervised classification with a simple Support Vector Machine (SVM) classifier.
\end{abstract}
\maketitle
\section{Introduction}
\label{sec:intro}
Hyperspectral Image (HSI) classification problem is a fundamental task for most applications in remote sensing areas. In recent years, this task has gained significant improvements from the deep learning technology for its powerful feature extraction ability. Some representative supervised feature extraction methods were developed for HSI classification task, such as deep convolutional neural networks (1D-CNN)\cite{hu2015deep}, deep recurrent neural networks (RNN)\cite{mou2017deep}, spectral-spatial residual network (SSRN)\cite{zhong2017spectral}, spectral-spatial multiscale feature extraction\cite{wang2020adaptive}, fully contextual network (FullyContNet)\cite{wang2021fully}, sparse and low-rank graph for discriminant analysis (SLGDA)\cite{li2016sparse}, supervised deep feature extraction (S-CNN)\cite{liu2017supervised}. The basic idea of these methods is to learn spatial and spectral features from adequate data with labels. An ideal training dataset is crucial to good classification performance.

However, these supervised learning methods have two main disadvantages for the HSI classification task. Firstly, supervised methods have high requirements for training datasets. Features learned by supervised methods may easily suffer from the distribution gap between training and testing samples, leading to a significant discrepancy between the training and testing accuracy. And it is not practical to provide adequate HSI data to solve this problem, where the collection of labeled HSI data is expensive. Secondly, supervised methods usually have poor generalization ability in HSI tasks. Generally, models trained by these supervised techniques are only applicable to their current labeled dataset and oriented task. So researchers started to move research focus into unsupervised learning for the HSI classification task.

\begin{figure}[t] 
\centering
\includegraphics[scale=0.3]{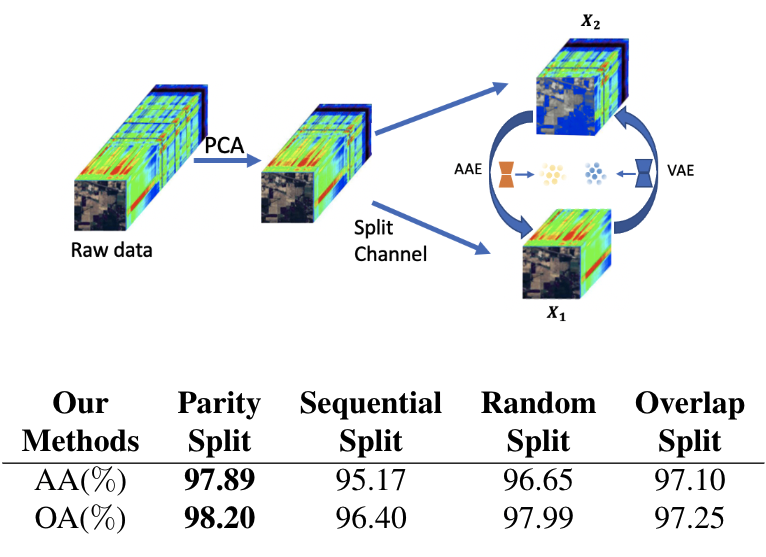}
\caption{Our cross-view-prediction strategy applies AAE and VAE to perform the cross channel prediction tasks to construct different views of original data for contrastive learning. Table in figure shows all of these methods achieve unsupervised state-of-the-art performance in IP dataset.}
\label{Cross}
\end{figure}

Ideally, unsupervised learning algorithms can address these disadvantages for the HSI data by extracting representative features from the data itself without labels, which is very efficient and practical. Traditional unsupervised representation methods such as principal component analysis (PCA)\cite{wold1987principal} and independent component analysis (ICA)\cite{villa2011hyperspectral} focus on reducing the dimension of input data, which ignores high-level information such as contextual semantics. Recent research uses unsupervised deep learning methods such as auto-encoder (AE) to perform representation learning to dig out features from the data itself. AE tries to map the input data into an embedding feature space and learn the distribution of samples. Famous AE-based methods for HSI classification tasks are stacked sparse autoencoder (SSAE)\cite{tao2015unsupervised} and three-dimension convolutional autoencoder (3DCAE)\cite{mei2019unsupervised}. Generally, these AE-based representative learning methods show an impressive feature extraction ability and classification performance.

In addition, contrastive learning is also a promising method for exploring the feature of data, which can learn the representation of data with high-level information. The basic rule of this method is to sample two data and learn contrastive features by minimizing the feature distance between the data of the same class and maximizing the feature distance between the data of different classes. Contrastive learning improves the performance of downstream tasks by discriminating different samples instead of estimating the true distribution of the samples. Typical contrastive learning methods are MOCO-V1\cite{chen2020improved}, MOCO-V2\cite{he2020momentum} and SimCLR\cite{chen2020simple}. Also, BYOL\cite{grill2020bootstrap} has a similar idea to get the state-of-art performance, which learns contrastive features via mutual prediction between different augmentation transformations without negative pairs. These unsupervised contrastive methods generally have a similar performance as fully-supervised methods in some image classification tasks.

Recent research in remote sensing areas gains many successes through combining representation learning and contrastive learning. Representation learning pays more attention to each image pixel but lacks semantic-level information. Oppositely, contrastive learning pays more attention to semantic information but ignores context details. By combining representation learning and contrastive learning, methods such as ContrastNet\cite{cao2021contrastnet}, Vision transformer-based contrastive learning\cite{hu2021contrastive}, Spatial-Spectral Clustering methods\cite{hu2021deep} show a remarkable potential to reduce dependency on datasets and achieve state-of-the-art performance. 

Unlike previous methods, we propose a self-supervised hyperspectral feature learning method based on cross-representation learning and contrastive learning. The whole algorithm is a cross-view-prediction style (Figure \ref{Cross}). Inspired by Split-AE\cite{zhang2017split}, and considering the characteristic of hyperspectral images, we propose a cross-representation learning method to induce contrastive features. In particular, hyperspectral images share redundant spectral-spatial features and preserve high-level semantics across the channel, which naturally leads to cross-channel prediction tasks. To exploit the consistency of semantic information between channels, we split the input data into two subsets and force networks to predict one subset of the data from the other. The resulting latent codes can be seen as different views of original data and can induce better features by contrastive learning methods. We try to explore consistently semantic features by this cross-view-prediction style. Besides, this style makes feature transfer well between different sub-spaces, which explores a better representation for other unseen downstream tasks.

One novelty here is using cross-channel-prediction tasks as augmentation methods to construct different descriptions of original data, encouraging exploring contrastive features. Another novelty is treating cross-representation latent codes as different views and using a contrastive learning method to generate a better representation feature. The main idea of our contrastive learning method is to maximize the mutual information and minimize conditional entropy between two views. The pipeline is interpretable and straightforward without a requirement for negative samples. 

Benefiting from our cross-view-prediction and contrastive learning strategy, we extract better features for HSI classification tasks and get a state-of-the-art performance by simply connecting a Support Vector Machine (SVM)\cite{steinwart2008support}. To summarize, our contribution can be listed below:
\begin{itemize}
\setlength\itemsep{-.3em}
\item We propose a cross representation learning method exploring high dimension HSI data. Learning from cross-channel prediction tasks, the model implicitly learns contrastive latent codes as augmentations in hyperspectral space.
\item We propose a self-supervised contrastive feature learning method for the HSI classification task. Specifically, the proposed method views the cross-representation latent codes as different views of original data and optimizes the features. We achieved state-of-the-art hyperspectral imagery classification performance only with the help of a weak classifier SVM in several standard datasets.
\end{itemize}

\section{Related work}
\subsection{Autoencoder based feature learning}
Autoencoder (AE) is commonly used in unsupervised learning and representation learning structure. Traditional autoencoder has no constraints over latent space. VAE and AAE is then proposed to optimize the distribution of the latent code, which makes latent code more representative and disentangled. Variational autoencoder(VAE)\cite{kingma2013auto} uses Kullback–Leibler (KL) divergence and reparameterization method to set latent code into norm distribution. Also, a reconstruction phase ensures the distribution of generated image and input image to be close. Adversary autoencoder (AAE)\cite{makhzani2015adversarial} uses a  generative adversarial network (GAN) style to set adversary restriction over latent code. The encoder of AAE plays the role of a generator to generate the latent code into target distribution and fool discriminator. The discriminator is trained to distinguish the randomly sampled norm distribution and latent code. Besides, a reconstruction phase also ensures the distribution of generated image and input image to be close. Some past research\cite{cao2021contrastnet}\cite{8989966}\cite{9109663} on representation learning for hyperspectral images shows the latent codes generated by VAE and AAE are relatively fixed and better representative. So VAE and AAE are widely used in hyperspectral tasks for better classification performance.

\subsection{Unsupervised feature learning}
ContrastNet\cite{cao2021contrastnet} treats different encoders as augmentation functions, and uses the discriminative learning method (prototypical contrastive learning) to fuse different feature spaces. This method shows the potential of getting more representative features by applying contrastive learning over different feature spaces. Different from ContrastNet, we apply a cross representation learning strategy to induce a contrastive feature and treat cross representations as different `views' of original data to optimize. Besides, compared with the complex M-N steps optimizing process, our pipeline has no requirements for negative data, which is more straightforward.

Split-brain-autoencoder (Split-AE)\cite{zhang2017split} is a typical cross-channel prediction unsupervised pipeline. Split-AE trains two disjoint sub-networks to predict one subset of raw data from another subset of raw data. 
Training these cross-channel encoders to predict `unseen' channels makes Split-AE implicitly learn contrastive features for other tasks. Inspired by Split-AE\cite{zhang2017split}, a cross representation learning strategy is naturally designed for the hyperspectral image because of its high channel dimension. In this way, we explore contrastive features and construct different descriptions of the raw data for the later contrastive learning stage.

Bootstrap Your Own Latent (BYOL)\cite{grill2020bootstrap} is a self-supervised method without the requirement of negative data. It uses one online and one target network with the same structure to learn from each other and applies a moving average update strategy. Some other self-supervised methods, such as  SimSiam \cite{chen2021exploring} and PixContrast\cite{xie2021propagate}, also use similar ideas. Inspired by this, we use this style to optimize our contrastive multi-descriptions task. Because of the exponential moving average strategy and information entropy-based loss, our method can avoid trivial solutions. Also, removing the need for negative samples makes the pipeline efficient and straightforward.

\section{Proposed Method}
Our cross-view-prediction style can be divided into two stages: cross representation learning stage and contrastive learning stage (Algorithm \ref{two_stage}). $\mathcal{Z}_{aae}$,$\mathcal{Z}_{vae}$,$\mathcal{Z}_{con}$ are features extracted by AAE, VAE and contrastive net. $\mathcal{H}$ and $\mathcal{P}$ represent cross-representation and contrastive learning process. In the first stage, we trained VAE and AAE to learn a pair of representations of original data from cross-channel-prediction tasks. In the second stage, we use a contrastive network to optimize representation features. 

\begin{algorithm}[tb] 
\caption{Cross-View-prediction Algorithm}
\textbf{Stage 1: Cross representation learning} \\
\begin{algorithmic}[1]
\State \texttt{$\mathcal{Z}_{vae},\mathcal{Z}_{aae}=\mathcal{H}(X_{raw})$}
\Comment{learn cross-representation}
\end{algorithmic}
\textbf{Stage 2: Contrastive learning} \\
\vspace{-10pt}
\begin{algorithmic}[1]
\State \texttt{$\mathcal{Z}_{con}=\mathcal{P}(\mathcal{Z}_{vae}, \mathcal{Z}_{aae})$}
\Comment{learn contrastive features}
\end{algorithmic}
\label{two_stage}
\end{algorithm}

\subsection{Cross representation learning}
One typical characteristic of our method is using a cross-representation learning style. We try to explore the contrastive features from hyperspectral data itself. Considering the consistency of semantic information between adjacent channels, we split the input data into two subsets along the channel direction and force networks to predict one subset of the data from the another. Ideally, there is no consistent information loss in this cross prediction task from the original data\cite{zhang2017split}. 

We propose four cross representation learning methods for hyperspectral self-supervised learning tasks to get different views that preserve consistent information. They are parity split, sequential split, random split, overlapping split. Intuitively, we see these learning methods as augmentation methods and treat resulting latent codes as different views with consistent semantics. 
The pipeline of these four augmentation methods is shown in Figure \ref{Cross} and Algorithm \ref{alg:Cross}. $X[:,:,:,0:C:2]$ means taking channels with index from $0$ to $C$ with step $2$. $:$ means no operation along this dimension.


\begin{algorithm}[tb] 
\caption{Cross representation learning for HSI}
\textbf{Input:} \\
Raw hypersepctral image:  $X_{raw}$\\
VAE encoder, AAE encoder: $\mathcal{F}_{\eta E}$, $\mathcal{F}_{\mu E}$\\ 
\vspace{-13pt}
\begin{algorithmic}[1]
  \For{each batch}
    \State \texttt{$X=PCA(X_{raw})$}
    \State \texttt{$X \in \mathbb{R}^{H \times W \times C}$} \Comment{apply PCA into original data}
    \If {`Parity split'}
    \State{
    $X_{1}=X[:,:,0:C:2]$,
    $X_{2}=X[:,:,1:C:2]$
    }
    \ElsIf {`Sequential split'}
    \State{
    $X_{1}=X[:,:,0:\frac{C}{2}], X_{2}=X[:,:,\frac{C}{2}:C]$
    }
    \ElsIf {`Random split'}
    \State{
    $X_{rs}=\mathcal{RS}_{\frac{C}{6}}(\frac{C}{2}:C), X_{2}=X[:,:,\frac{C}{3}:C]$}
    \State{
    $X_{1}=Cat(X[:,:,0:\frac{C}{2}],X_{rs})$}
    \ElsIf {`Overlapping split'}
    \State{
    $X_{1}=X[:,:,0:\frac{2}{3}C], X_{2}=X[:,:,\frac{1}{3}C:C]$
    }
    \EndIf
    \State
    $\mathcal{Z}_{vae}=\mathcal{F}_{\eta E}(X_1)$    
    \Comment{Train VAE representation}
    
    \State
    $\mathcal{Z}_{aae}=\mathcal{F}_{\mu E}(X_2)$
    \Comment{Train AAE representation}
  \EndFor
\end{algorithmic}
    \vspace{-6pt}
    \textbf{Output:} VAE and AAE representation view: $\mathcal{Z}_{vae}$, $\mathcal{Z}_{aae}$
    \label{alg:Cross}
\end{algorithm}

We trained AAE and VAE separately because their latent codes are similar but slightly different, and these two structures work best in remote sensing areas\cite{cao2021contrastnet}. To process hypersepctral image, we first apply the Principal component analysis (PCA) algorithm\cite{wold1987principal} as pre-process to reduce the channel numbers. Then we use sliding window method to split the whole HSI into small patches, geting the data $X \in \mathbb{R}^{H \times W \times C}$. We then apply our augmentation methods to split the data $X$ into two parts $X_1\in \mathbb{R}^{H \times W \times C_1}$, $X_2\in \mathbb{R}^{H \times W \times C_2}$. $H, W$ are patch size we created for our dataset. To retain all information from original data, $X_1$ and $X_2$ should loop through all data $X \subseteq (X_1 \cup X_2)$.

For `Parity split': We use odd channels subset as $X_1$, even channels subset as $X_2$. $X_1$ and $X_2$ are designed as adjacent channel styles to exploit robust consistency.

For `Sequential split': We use the first-half channels as $X_1$ and the left as $X_2$. $X_1$ and $X_2$ are designed as one after the other style to exploit more long-term consistency.

For `Random split': $X_1$ is stacked by a fix $\frac{C}{2}$ channels part and $\frac{C}{6}$ random channels part along channel dimension. $X_2$ is fixed when working as prediction ground truth. $\mathcal{RS}_{\frac{C}{6}}(\frac{C}{2}:C)$ means random sampling $\frac{C}{6}$ channels from index $\frac{C}{2}$ to index $C$. $Cat$ means concatenation operation along channel dimension. By introducing extra random information, we want to make the strategy explore extra more information on its consistency.

For `Overlap split': We use first $\frac{2}{3}$ channels as $X_1$, second $\frac{2}{3}$ channels as $X_2$. $X_1$ and $X_2$ are designed overlapping to encourage trade-off between long term and short term consistency.

After split augmentation, we got two views $X_1$ and $X_2$ for VAE and AAE. We trained a VAE model $\mathcal{F}_1$ with data $X_1$ to predict $X_2$, $\hat{X_2}=\mathcal{F}_1(X_1)$. Symmetrically, we use same strategy to train a AAE model $\mathcal{F}_2$ with data $X_2$ to predict $X_1$, $\hat{X_1}=\mathcal{F}_2(X_2)$. The losses used in this training process can be mean squared error loss (Equation \ref{loss_mse})or standard cross-entropy. $W,H,C$ is the size of the images $X,Y$.
\begin{equation}
    \mathcal{L}_{mse}(X,Y)=\frac{1}{W*H*C}\sum_{h,w,c} \lVert X-Y\rVert^2
    \label{loss_mse}
\end{equation}

For fair comparison without label introduced, we use mean square loss $\mathcal{L}_{mse}(\mathcal{F}_1(X_1),X_2)$, and $\mathcal{L}_{mse}(X_1,\mathcal{F}_2(X_2))$ in our VAE and AAE training.

\begin{figure}[tb] 
\centering
\includegraphics[scale=0.31]{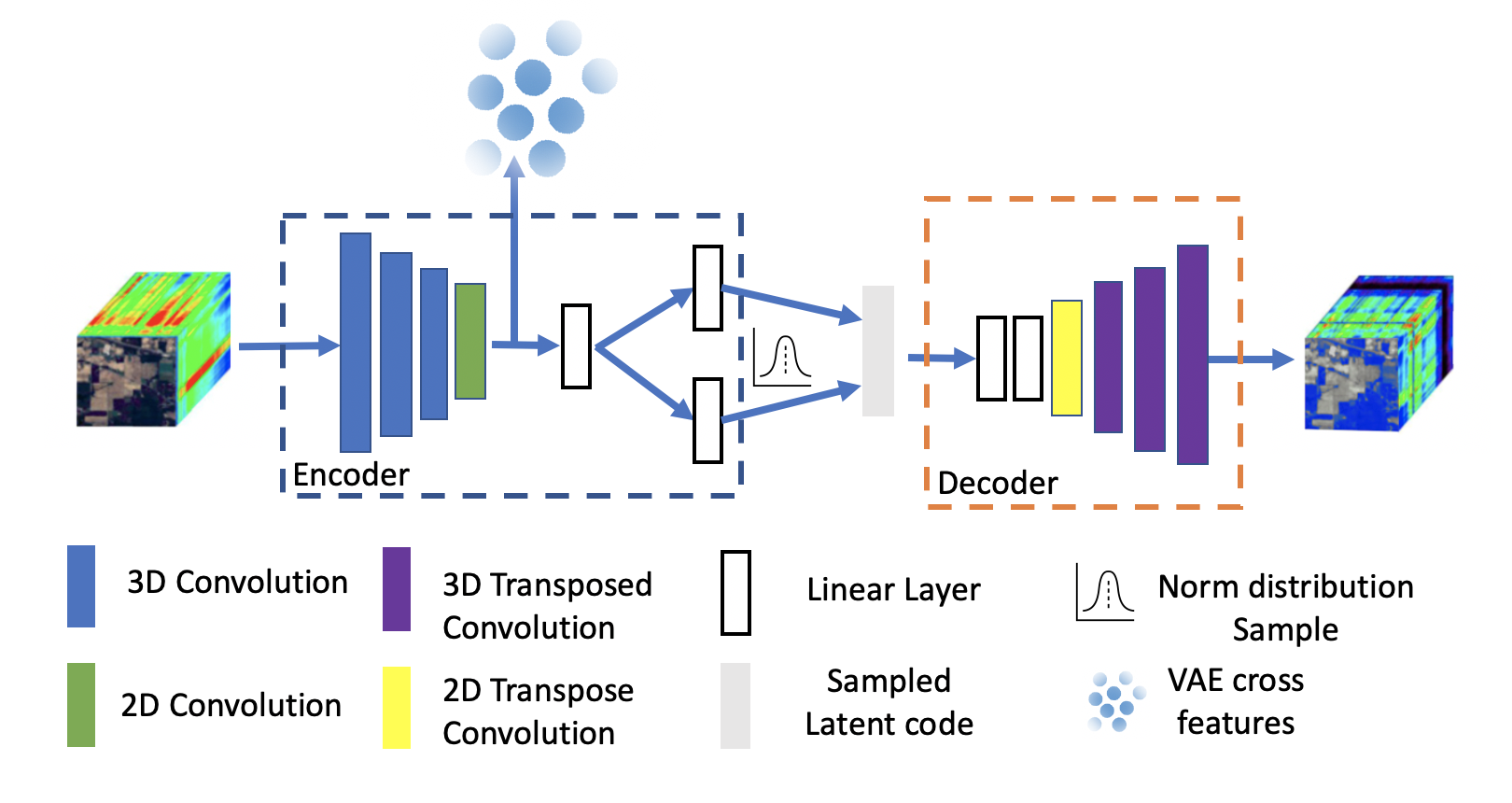}
\caption{VAE structure for hypersepctral image}
\label{VAE_pipeline}
\end{figure}
\subsubsection{VAE module}
To perform this cross channel prediction task $\hat{X}_2=\mathcal{F}_1(X_1)$, we designed a VAE feature extractor. The structure of this extractor is shown in Figure \ref{VAE_pipeline}. 

In Encoder, we used 3D and 2D CNN to generate latent code, learning more channel information with low computation complexity. Similar to the original VAE\cite{kingma2013auto}, two FC layers mapped features into mean $\mu$ and variance $\sigma$. We generated latent code by reparametering a norm distribution (Equation \ref{vae_code}), where $X$ is sampled from a norm distribution.

\begin{equation}
z=\mu+\sigma \times X, X \sim N(0,1)
\label{vae_code}
\end{equation}

In Decoder, symmetrical to Encoder, several FC layers, 2D and 3D transposed CNNs are used to recover the inputs. Loss function of VAE is Equation \ref{loss_vae}. 

\begin{equation}
\begin{array}{lc}
    \mathcal{L}_{vae}=\mathcal{L}_{mse}^{vae}+L_{kl}\\[1mm]
    \mathcal{L}_{mse}^{vae}=\mathcal{L}_{mse}(\mathcal{F}_1(X_1),X_2)\\[1mm]
    \mathcal{L}_{kl}=\frac{1}{2}\sum_{i}^N(\mu_i^2+\sigma_i^2-log\sigma_i^2-1)
\end{array}
    \label{loss_vae}
\end{equation}

$\mathcal{L}_{mse}^{vae}$ is mean square error loss, ensuring the reconstruction performance. $\mathcal{F}({X}_1)$. $L_{kl}$ is distribution loss, minimizing the KL divergence between latent code and normal distribution. 

\begin{figure}[tb] 
\centering
\includegraphics[scale=0.31]{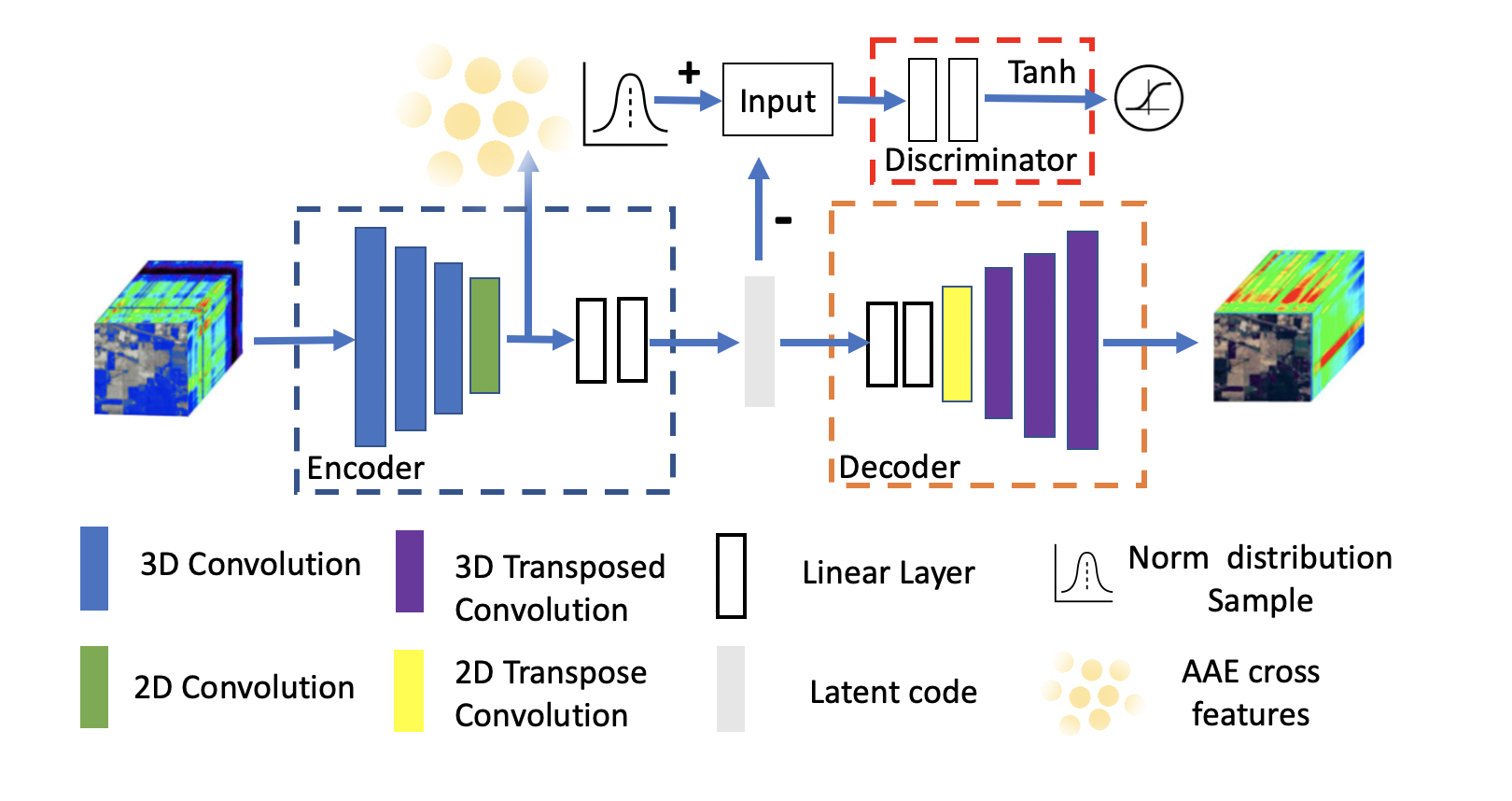}
\caption{AAE structure for hypersepctral image}
\label{AAE_pipeline}
\end{figure}
\subsubsection{AAE module}
Intuitively, we want to design two feature extractors in the Siamese style. So we also design an AAE module siamese to VAE to perform the prediction $\hat{X}_1=\mathcal{F}_2(X_2)$. The structure of this feature extractor is shown in Figure \ref{AAE_pipeline}.

The encoder and decoder have the same structure as the previous VAE. The different part is encoder here also plays a role of GAN generator. Similar to the original AAE\cite{makhzani2015adversarial}, we feed the latent code and a random norm distribution sample into a discriminator. There are two-phase in AAE training: the reconstruction phase and the regularization phase. In reconstruction phase, we train the encoder and decoder with loss $L_{mse}^{aae}$ in Equation \ref{loss_mse_aae}. 
\begin{equation}
    \mathcal{L}_{mse}^{aae}=\mathcal{L}_{mse}(\mathcal{F}_2(X_2),X_1)
    \label{loss_mse_aae}
\end{equation}

\begin{figure*}[t] 
\centering
\includegraphics[scale=0.45]{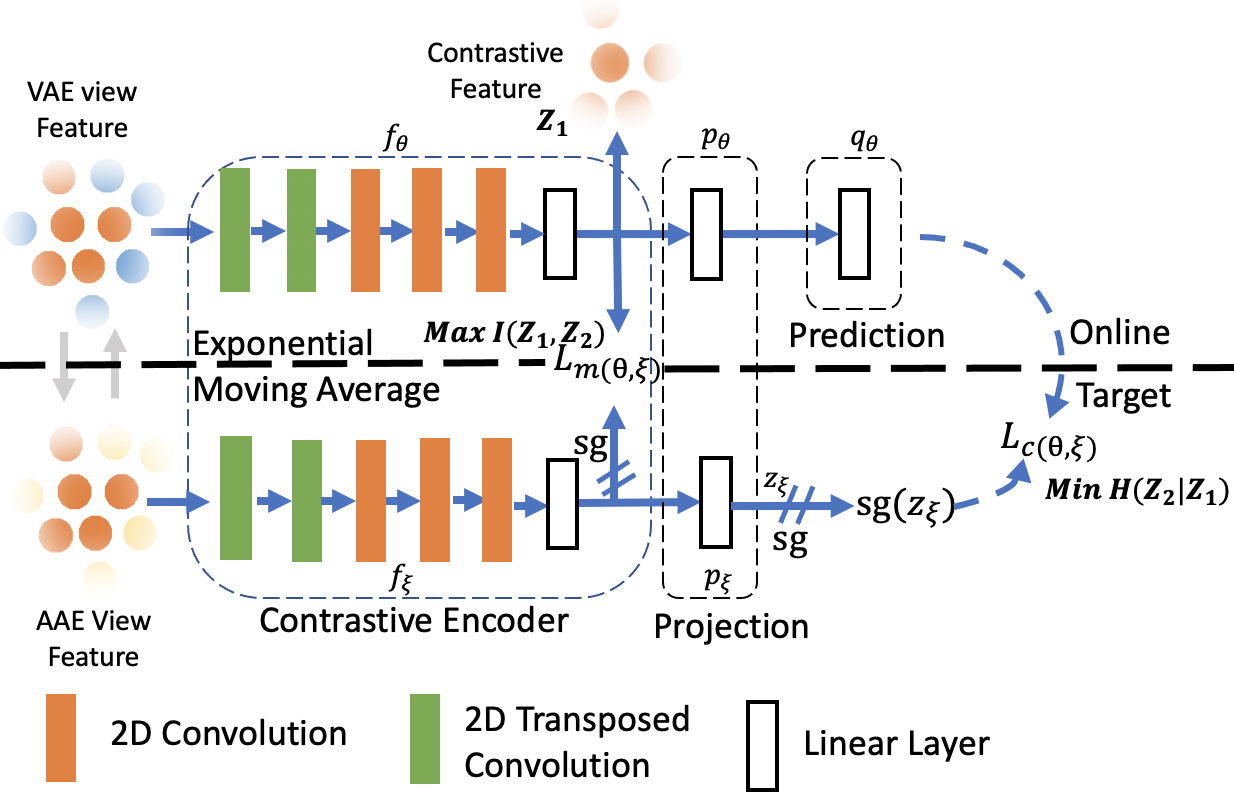}
\caption{Structure of the system, sg means stop-gradient, $I(Z_1,Z_2)$ means mutual information between lantent $Z_1$ and $Z_2$,  $H(X_2|X_1)$ means conditional entropy}
\label{Model_pipeline}
\end{figure*}

In the regularization phase, we optimize the discriminator first and then the generator. For the loss, we use the Wasserstein GAN (WGAN)\cite{arjovsky2017wasserstein} loss. $\mathcal{L}_{G}$ is generator loss.
\begin{equation}
    \mathcal{L}_{G}=E_{x\sim P_g}[D(x)]-E_{x\sim P_r}[D(x)]
\label{loss_aae_G}
\end{equation}
\begin{equation}
    \mathcal{L}_{D}=-E_{x\sim P_g}[D(x)]
    \label{loss_aae_D}
\end{equation}

$\mathcal{L}_{D}$ is discriminator loss. $D(x)$ means the output of the discriminator with input $x$.  $x\sim P_g$ means sampling input from distribution $P_g$. $P_r$ is the distribution of real samples. $P_g$ is the distribution of fake samples.

\subsection{Contrastive learning}
Another typical characteristic of our method is that we treat mutual prediction features as different views of the original data and design a contrastive learning method to optimize the features further.

Figure \ref{Model_pipeline} is an overview of our contrastive learning method. We feed the VAE representation view and AAE representation view into the pipeline (both are treated as online net once and target network once symmetrically). Our method consists of Contrastive Encoder module ($f_{\theta},f_{\xi}$), Projection module ($p_{\theta},p_{\xi}$) and Prediction module $q_\theta$, which all share the same structure for online and target nets. 

Because our cross representation naturally learned some good contrastive features, our Contrastive Encoder module is simple with several 2D CNN and 2D Transpose CNN. The projection and prediction module consists of several FC layers, which is similar to BYOL\cite{grill2020bootstrap}, SimSiam\cite{chen2021exploring}. 

The detailed algorithm is shown in Algorithm \ref{alg:BYOL}. We use Contrastive Encoder module to encode input features and induce a more representative contrastive feature. We use two information entropy strategies to promote this purpose: maximizing mutual information and minimizing the conditional entropy between input features, which is simple and interpretable. 

\begin{algorithm}[htbp] 
\caption{Contrastive learning for HSI}
\textbf{Input:} \\
 VAE and AAE representation view: $\mathcal{F}_{vae}$, $\mathcal{F}_{aae}$\\
    Online encoder, projector, predictor: $f_\theta$, $p_{\theta}$, $q_{\theta}$\\ 
    Target encoder, projector, Optimizer:$f_{\xi}$, $p_{\xi}$, $Opt$\\
\vspace{-13pt}
\begin{algorithmic}[1]
  \For{each batch}
    \State \texttt{$\mathcal{F}_{vae}$, $\mathcal{F}_{aae}$} \Comment{ Sample a pair of features}
    \State \texttt{$f_1=f_{\theta}(\mathcal{F}_{vae})$, $f_2=f_{\theta}(\mathcal{F}_{aae})$} \Comment{ Online feature}
    \State \texttt{$f_1'=f_{\xi}(\mathcal{F}_{vae})$, $f_2'=f_{\xi}(\mathcal{F}_{aae})$} \Comment{ Target feature}
    \State \texttt{$\mathcal{L}_{m}(Z_1,Z_2)=-(I(Z_1,Z_2)+\alpha (H(Z_1)+H(Z_2)))$}
    \State
    \Comment{Mutual Loss}
    \State \texttt{$z_1= p_{\theta}(f_1)$}, \texttt{$z_2= p_{\theta}(f_2)$} \Comment{ Online projection}
    \State \texttt{$z_1' = p_{\xi}(f_1')$, $z_2' = p_{\xi}(f_2')$} \Comment{ Target projection}
    \State \texttt{$\mathcal{L}_{c}=-2(\frac{<q_\theta(z_1),z_1'>}{\lVert q_{\theta}(z_1) \rVert \lVert z_1'\rVert}+\frac{<q_\theta(z_2),z_2'>}{\lVert q_{\theta}(z_2) \rVert \lVert z_2'\rVert})$} 
    \State
    \Comment{Conditional loss}
    \State \texttt{$\mathcal{L} =\lambda \mathcal{L}_m+ \mathcal{L}_c$} \Comment{Loss function}
    \State \texttt{$\theta \gets Opt(\theta,\nabla_\theta \mathcal{L})$} \Comment{Update online network}
     \State \texttt{$\xi \gets \tau\xi +(1-\tau)\theta$} \Comment{Update target network}
  \EndFor
\end{algorithmic}
\vspace{-5pt}
    \textbf{Output:} Online encoder: $f_\theta$
\label{alg:BYOL}
\end{algorithm}
\begin{table}[h]
\caption{Details of IP dataset}
\label{tab:IP_data_tabel}
\begin{tabular*}{\columnwidth}{ccc}
\multicolumn{3}{c}{\textbf{IP dataset}} \\
\hline
\textbf{Class} & \textbf{Index}& \textbf{Sample number}\\
\hline
Alfalfa & 0& 46\\
Corn-notill & 1& 1428\\
Corn-mintill & 2& 830\\
Corn & 3& 237\\
Grass-pasture & 4& 483\\
Grass-trees & 5& 730\\
Grass-pasture-mowed & 6& 28\\
Hay-windrowed & 7& 478\\
Oats & 8& 20\\
Soybean-notill & 9& 972\\
Soybean-mintill & 10& 2455\\
Soybean-clean & 11& 593\\
Wheat & 12& 205\\
Woods & 13& 1265\\
Building-Grass-Tree-Drive & 14& 386\\
Stone-Steel-Towers & 15& 93\\
\hline
\textbf{Total} & &\textbf{10249}\\
\hline
\end{tabular*}
\end{table}

\subsubsection{Maximize mutual information}
Instead of maximizing the lower bound of mutual information, we use a mutual-information-based loss (Equation \ref{loss_mutal}) to optimize the mutual information directly. $I(Z_1,Z_2)$ means mutual information between $Z_1,Z_2$. $H$ is the information entropy. $\alpha$ is a regularization parameter. With this loss $\mathcal{L}_{m}$, we maximize mutual information between $Z_1$ and $Z_2$. Also, we try to maximize the information entropy of $Z_1$ and $Z_2$ to preserve more information for each view, which avoids trivial solutions. 
\begin{equation}
    \mathcal{L}_{m}(Z_1,Z_2)=-(I(Z_1,Z_2)+\alpha (H(Z_1)+H(Z_2)))
\label{loss_mutal}
\end{equation}

To compute the $I(Z_1,Z_2)$ and $H(Z_i)$, we define the joint probability distribution $P(z_1,z_2)\in ({d,d})$ in Equation \ref{joint_P}. 
\begin{equation}
    P(z_1,z_2)=\mathcal{S}(Z_1)\mathcal{S}(Z_2)^T
\label{joint_P}
\end{equation}

$ P(z_1,z_2)$ shows the correlation between two random variables of representations. $\mathcal{S}$ is Softmax operation. And then we could compute the Equation \ref{loss_mutal} by Equation \ref{loss_mutal_p}.

\begin{equation}
    \mathcal{L}_{m}=- \sum_{i=1}^d \sum_{j=1}^d P(i,j) \ln \frac{P(i,j)}{P^{\alpha+1}_i P^{\alpha+1}_{j}}
\label{loss_mutal_p}
\end{equation}

$d$ is the dimension of the our representation feature. $\alpha$ is a normalization parameter for the loss. 

\subsubsection{Minimize conditional entropy}
Our network uses a prediction operation after the Projection module to minimize the conditional entropy, which encourages maximizing the overlapping information over different views. As we know, when two views are fully overlapped, then $H(Z_2|Z_1)=0$, which means the resulting feature is most representative. This design can encourage contrastive encoder to encode more representative information\cite{grill2020bootstrap}\cite{chen2021exploring}.

To reduce conditional entropy, common methods is to introduce a variational Gaussian distribution\cite{goodfellow2014generative}. In practice, this minimization problem can also be converted into Equation \ref{convert}\cite{lin2021completer}, where $G^i$ means a mapping between views. Our network (Figure \ref{Model_pipeline}) and loss $ \mathcal{L}_{c}$ (Equation \ref{loss_condition}) is designed based on this, where we compute the distance between the prediction output $q_{\theta}(z_i)$ and $z_i'$ with l2-normalization.
\begin{equation}
    \min \lVert Z_1-G^1(Z_2) \rVert^2+\lVert Z_2-G^2(Z_1) \rVert^2
    \label{convert}
\end{equation}
\begin{equation}
    \mathcal{L}_{c}=2-2(\frac{<q_\theta(z_1),z_1'>}{\lVert q_{\theta}(z_1) \rVert \lVert z_1'\rVert}+\frac{<q_\theta(z_2),z_2'>}{\lVert q_{\theta}(z_2) \rVert \lVert z_2'\rVert})
\label{loss_condition}
\end{equation}

$z_1$ and $z_1'$ are extracted contrastive features from VAE and AAE views, the same as the $z_2$ and $z_2'$. $q_\theta$ is a prediction operation. 

\subsubsection{Update strategy}
We symmetrize the training process by exchanging VAE cross-feature and AAE cross-feature into online and target net. We use the weighted loss (Equation \ref{total_loss}) to optimize the online net by backpropagation and use the exponential moving average strategy to update the target net (Equation \ref{update_online},\ref{update_target}). 
\begin{equation}
   \mathcal{L} =\lambda \mathcal{L}_m+ \mathcal{L}_c
   \label{total_loss}
\end{equation}

\begin{equation}
   \theta \gets Opt(\theta,\nabla_\theta L)\\[1mm]
    \label{update_online}
\end{equation}

\begin{equation}
    \xi \gets \tau\xi +(1-\tau)\theta   
    \label{update_target}
\end{equation}

$\lambda$ is the loss weight. $\mathcal{L}_m,\mathcal{L}_c$ can be computed by Equation (\ref{loss_mutal_p}, \ref{loss_condition}). $\tau$ is a target delay rate. This updating strategy can avoid some trivial solutions. $Opt$ is the optimizer.

\begin{table}[h]
\caption{Details of PU ataset}
\begin{tabular*}{\columnwidth}{ccc}
\multicolumn{3}{c}{\textbf{PU dataset}} \\
\hline
\textbf{Class} & \textbf{Index}& \textbf{Sample number}\\
\hline
Asphalt & 0& 6631\\
Meadows & 1& 18649\\
Gravel & 2& 2099\\
Trees & 3& 3064\\
Painted metal sheets & 4& 1345\\
Bare Soil & 5& 5029\\
Bitumen & 6& 1330\\
Self-Blocking-Bricks & 7& 3682\\
Shadows&8&947\\
\hline
\textbf{Total} & &\textbf{42776}\\
\hline
\end{tabular*}
\label{PU_data_tabel}
\caption{Details of PU ataset}
\end{table}
\begin{table}[h]
\begin{tabular*}{\columnwidth}{ccc}
\multicolumn{3}{c}{\textbf{SA dataset}} \\
\hline
\textbf{Class} & \textbf{Index}& \textbf{Sample number}\\
\hline
Brocoli-green-weeds1 & 0& 2009\\
Brocoli-green-weeds2 & 1& 3726\\
Fallow & 2& 1976\\
Fallow-rough-plow & 3& 1394\\
Fallow-smooth& 4& 2678\\
Stubble& 5& 3959\\
Celery& 6& 3579\\
Grapes-untrained& 7& 11271\\
Soil-vinyard-develop & 8& 6203\\
Corn-senesced-green-weeds& 9& 3278\\
Lettuce-romaine-4wk& 10& 1068\\
Lettuce-romaine-5wk& 11& 1927\\
Lettuce-romaine-6wk & 12& 916\\
Lettuce-romaine-7wk& 13& 1070\\
Vinyard-untrained & 14& 7268\\
Vinyard-vertical-trellis & 15& 1807\\
\hline
\textbf{Total} & &\textbf{54129}\\
\hline
\end{tabular*}
\caption{Details of SA dataset}
\label{SA_data_tabel}
\end{table}
\section{Experiments and analysis}
We use a simple classifier SVM to achieve the state-of-the-art classification performance to show the representation ability of our extracted features.
\subsection{Datasets}
We apply our method to three standard datasets for the hyperspectral classification task:  Indian Pines(IP) dataset, University of Pavia(PU) dataset, and Salinas Scene(SA) dataset.

IP dataset consists of $145 \times 145$ pixels with  $220$ spectral reflectance bands in the wavelength range $0.4-2.5\times10^{-6}m$. There are 16 classes of ground truth.  Without the bands covering water absorption, 200 bands were considered in the experiments. The main challenge in this dataset is an unbalanced number of labels. Some class likes Soybean-mintill has 2455 samples, but class like Oats only has 20 samples. The detail can be seen in the table \ref{tab:IP_data_tabel} below. IP dataset (16 classes in ground truth) consists of $145 \times 145$ pixels with  $220$ spectral reflectance bands. The main challenge in the IP dataset is the unbalanced number of labels. 

\begin{table*}[t]
\caption{Evaluation in Indian Pines dataset}
\begin{tabular*}{\textwidth}{@{\extracolsep{\fill}}ccccccccc}
\hline
\textbf{Class}&\multicolumn{3}{c}{\textbf{Supervised Feature Extraction}}&\multicolumn{5}{c}{\textbf{Unupervised Feature Extraction}} \\
\hline
\textbf{ } & \textbf{SLGDA}& \textbf{1D-CNN}& \textbf{S-CNN}& \textbf{3DCAE}& \textbf{AAE}& \textbf{VAE}& \textbf{ContrastNet}&\textbf{Ours model} \\
\hline
0 & 39.62& 43.33& 83.33& 90.48& \textbf{100.00}& \textbf{100.00}& 85.37&92.35 \\
1 & 85.56& 73.13& 81.41& 92.49& 81.63& 78.78& \textbf{97.15}&96.95 \\
2 & 74.82& 65.52& 74.02& 90.37& 95.27& 92.37& 97.95&\textbf{97.86} \\
3 & 49.32& 51.31& 71.49& 86.90& \textbf{99.22}& 97.34& 95.62&93.64 \\
4 & 95.35& 87.70& 90.11& 94.25& 95.17& 96.09& 94.82&\textbf{98.40} \\
5 & 95.59& 95.10& 94.06& 97.07& 98.73& 98.27& 96.80&\textbf{99.40}\\
6 & 36.92& 56.92& 84.61& 91.26& 96.00& 98.67& 70.67&\textbf{100.00}\\
7 &99.75& 96.64& 98.37& 97.79& 99.84& \textbf{99.77}& 98.68&99.07\\
8 & 5.00& 28.89 & 33.33& 75.91& 96.30& 98.15& 70.37&\textbf{100.00}\\
9 & 69.11& 75.12& 86.05& 87.34& 87.01& 78.86& 97.45&\textbf{97.71}\\
10 & 89.91& 83.49& 82.98&90.24& 89.08& 81.75& 98.40&\textbf{98.73}\\
11 & 86.78& 67.55& 73.40&\textbf{95.76}& 93.51& 90.64& 93.57&94.19\\
12 & \textbf{99.51}& 96.86& 87.02&97.49& 98.56 &98.56& 95.32&98.92\\
13 & 96.45& 96.51& 94.38&96.03& 95.73& 93.24& 98.51&\textbf{99.91}\\
14 & 61.79& 39.08& 75.57&90.48& 97.31& 97.02& 96.73&\textbf{99.14}\\
15 & 84.16& 89.40& 79.76&98.82& 98.02& 98.81& 79.76&\textbf{100.00}\\
\hline
AA($\%$)& 73.10& 71.66& 84.44&92.04& 95.09& 93.51 &91.78&\textbf{97.89}\\
OA($\%$)& 85.19& 79.66& 80.72&92.35& 91.80& 88.03&97.08&\textbf{98.20}\\
\hline
\end{tabular*}
\vspace{-5pt}
\label{tabIP}
\end{table*}
\begin{table*}[tb]
\caption{Evaluation in Salinas dataset}
\begin{center}
\begin{tabular*}{\textwidth}{@{\extracolsep{\fill}}ccccccccc}
\hline
\textbf{Class}&\multicolumn{3}{c}{\textbf{Supervised Feature Extraction}}&\multicolumn{5}{c}{\textbf{Unupervised Feature Extraction}} \\
\hline
\textbf{ } & \textbf{SLGDA}& \textbf{1D-CNN}& \textbf{S-CNN}& \textbf{3DCAE}& \textbf{AAE}& \textbf{VAE}& \textbf{ContrastNet}&\textbf{Ours model} \\
\hline
AA($\%$)& 96.01& 94.73& 97.39& 97.45 &98.73& 97.67 &99.48&\textbf{99.70}\\
OA($\%$)& 93.31& 91.30& 97.92&95.81& 97.10 &95.23& 99.60&\textbf{99.82}\\
\hline
\end{tabular*}
\vspace{-5pt}
\label{tabSA}
\end{center}
\end{table*}
\begin{table*}[t]
\caption{Evaluation in University of Pavia dataset}
\begin{center}
\begin{tabular*}{\textwidth}{@{\extracolsep{\fill}}ccccccccc}
\hline
\textbf{Class}&\multicolumn{3}{c}{\textbf{Supervised Feature Extraction}}&\multicolumn{5}{c}{\textbf{Unupervised Feature Extraction}} \\
\hline
\textbf{ } & \textbf{SLGDA}& \textbf{1D-CNN}& \textbf{S-CNN}& \textbf{3DCAE}& \textbf{AAE}& \textbf{VAE}& \textbf{ContrastNet}&\textbf{Ours model} \\
\hline
AA($\%$)& 91.86& 87.34& 94.75 &95.36& 98.03& 94.16 &98.83&\textbf{99.30}\\
OA($\%$)& 94.15& 89.99& 92.78& 95.39& 98.14& 94.53& 99.46 &\textbf{99.78}\\
\hline
\end{tabular*}
\vspace{-5pt}
\label{tabPU}
\end{center}
\end{table*}

PU dataset (9 classes) consists of $610 \times 340$ pixels with  $103$ spectral reflectance bands. The main challenges in PU dataset are its complex background and many discontinuous pixels. PU dataset consists of $610 \times 340$ pixels with  $103$ spectral reflectance bands in the wavelength range $0.43-0.86\times10^{-6}m$. There are nine classes in the ground truth. The main challenges in this dataset are its complex background and many discontinuous pixels. The detail can be seen in the table \ref{PU_data_tabel} below.

SA dataset (16 classes) consists of $512 \times 217$ pixels with $224$ spectral reflectance bands. This dataset is simple compared with the IP dataset and PU dataset. SA dataset consists of $512 \times 217$ pixels with $224$ spectral reflectance bands in the wavelength range $0.36-2.5\times10^{-6}m$. Ignoring the bands covering water absorption, 204 bands remain in the experiments. There are 16 classes in the ground truth. This dataset is simple compared with the IP dataset and PU dataset. The detail can be seen in the table \ref{SA_data_tabel} below.

\begin{figure*}[tb]
\centering
\includegraphics[scale=0.62]{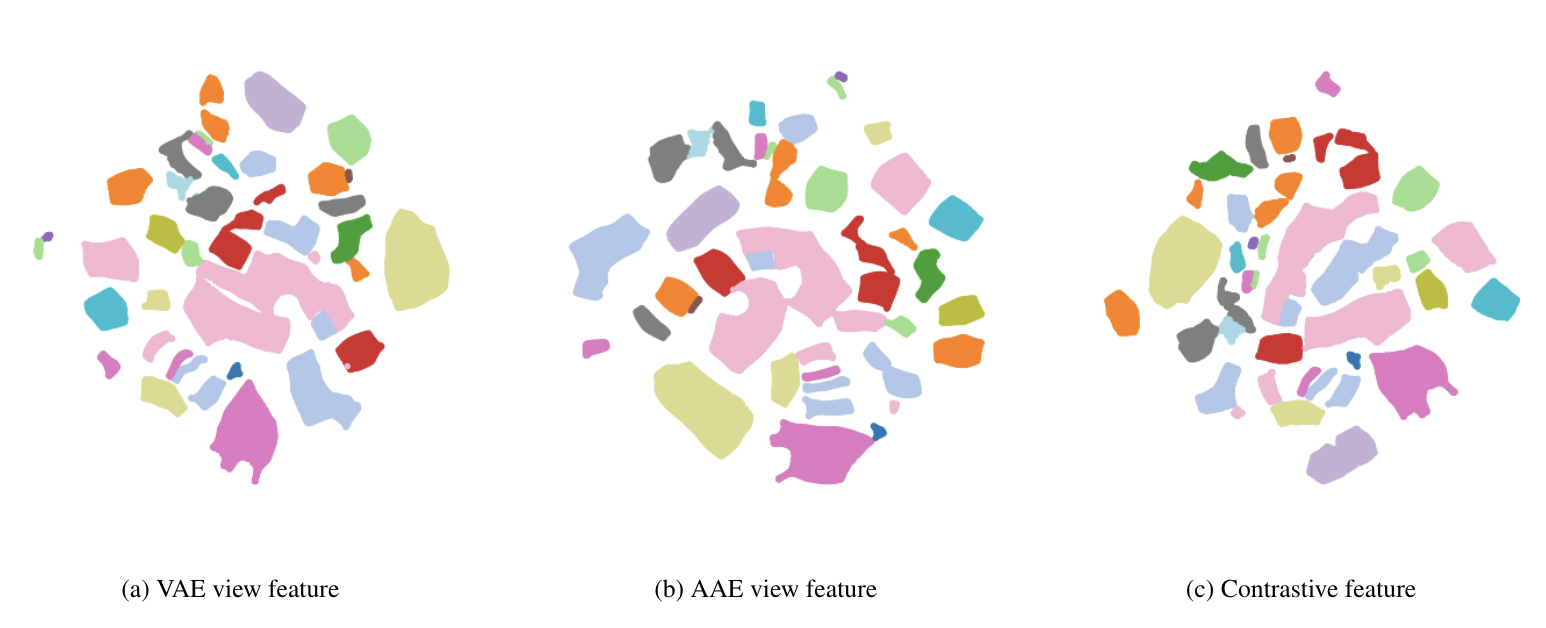}
\caption{t-SNE visualization of contrastive features in the IP dataset.}
\label{fig_tsne}
\end{figure*} 
\subsection{Model implementation}
To train SVM, we use $5\%$ samples in each class for training, and the left is used for testing in SA dataset (because the dataset is simple). However, in IP and PU datasets,  $10\%$ samples are used for training; the left is used for testing. The shape of patches after PCA is $(27,27,30)$. We do experiments on all four augmentation methods. Latent codes from VAE and AAE are $128$ dimension vectors. 

For AAE and VAE, Adam optimizers with a learning rate: 0.001 and weight decay: 0.0005. Two SGD optimizers for AAE has learning rate: 0.0001 (generator) and learning rate: 0.00005 (discriminator). The batch size is 256. We train the AAE module for 30 epochs and VAE for 40 epochs for the best performance. 

For contrastive learning, we use an Adam optimizer with a learning rate of $3e-4$, weight decay is set to 0.001. We train $200$ epochs and select the best feature. To normalize loss, $\alpha=9$, $\lambda=100$. Results are evaluated by average accuracy (AA) and overall accuracy (OA) (Equation \ref{Criterion}).
\begin{equation}
\begin{array}{lc}
OA(\%)=\frac{N_{p}}{N_{a}}\\[1mm]
AA(\%)=\frac{1}{N}\sum_{i=1}^{N}c_i\\[1mm]
\end{array}
 \label{Criterion}
\end{equation}

$N$ is class number, $c_i$ is each class classification accuracy. $N_{p}$ is the number of accurate classification samples. $N_a$ is the total number of all samples.
\subsection{Experiments Results}
\subsubsection{Cross representation experiments}
As PU and SA datasets are simple, all our strategies work very well. We compare our different augmentation methods in the IP dataset. Table in Figure \ref{Cross} shows HSI classification performance of our cross-view-prediction strategies. All four cross-prediction strategies have good performance, showing that our cross-view-prediction style encourages the consistency of semantic information and suits HSI tasks.

The `Parity Split' strategy works best, which is used to compare with other algorithms later. Random Split and Overlap Split also work state-of-the-art, which shows the advantage of our cross prediction style in finding robust and consistent information from random sampling. Sequential Split works a bit worse. It may be because our simple network has a limited ability to extract long-term consistency when faced with this simple split strategy.

Figure \ref{fig_tsne} shows the t-SNE visualization\cite{van2008visualizing} of the extracted features. Each color means one class. In our style, we could see AAE and VAE learned compact and discriminative features. It is hard to discriminate which distribution is better, but our contrastive learning method tries to fuse VAE and AAE feature distribution, which tends to make features more compact and informative.

\subsubsection{Contrastive learning experiments}
Table \ref{tabIP}, Table \ref{tabSA} and Table \ref{tabPU} show HSI classification experiments for IP, SA and PU dataset separately. Results in bold are the best performance. Some best performance are quoted from\cite{mei2019unsupervised}\cite{cao2021contrastnet}. Our model gets state-of-the-art performance with the highest OA and AA in all of three datasets. 

For the IP dataset, supervised feature learning methods have worse performance, which may result from the unbalanced dataset. However, our method doesn't work best in all classes. Contrastive learning method sometimes may confuse the feature representation for overlapping classes.

For the PU dataset, unsupervised feature learning methods outperform other supervised methods. We think it is because of its complex background, leading to trivial over-fitting results for supervised methods.

For the SA dataset, a simple dataset, our method gains $100\%$ results in most classes and still achieves the best accuracy. It shows our method indeed further extracts more information than the other unsupervised learning methods.

In summarization, unsupervised feature learning methods are better solutions for HSI classification tasks, whose labelled datasets are usually imbalanced and rare. Among unsupervised algorithms, our method gains the best performance and outperforms all other existing methods. This impressive result shows features from our cross-view-prediction style have a better representation ability. Also, results show our contrastive learning strategy further enhances the performance and optimizes the trade-off between the AAE feature view and the VAE feature view.

\section{Extra ablation experiments}
We have seen that our cross-view-prediction style has good performance in extracting HSI contrastive features. Cross-prediction and contrastive learning are two main strategies in our methods. In this section, we do the ablation experiments to check the effeteness of these two methods.
\begin{table}[t]
\caption{Ablation experiments of cross-prediction style}
\begin{tabular}{ccc|cc}
\hline
\textbf{ } & \textbf{VAE}& \textbf{Cross-VAE}& \textbf{AAE}& \textbf{Cross-AAE}\\
\hline
AA($\%$)& 88.65& \textbf{93.85}& 93.28&\textbf{95.71}\\
OA($\%$)& 82.02& \textbf{91.33}& 90.46&\textbf{92.13}\\
\hline
\end{tabular}
\begin{tablenotes}
    \footnotesize
    \item{Data in bold are the best performance.}
\end{tablenotes}
\label{Cross_ab}
\end{table}

\begin{figure*}[tbh]
\begin{minipage}[t]{0.24\linewidth}
\centering
\subfloat[Standard VAE view feature]{
\includegraphics[scale=0.4]{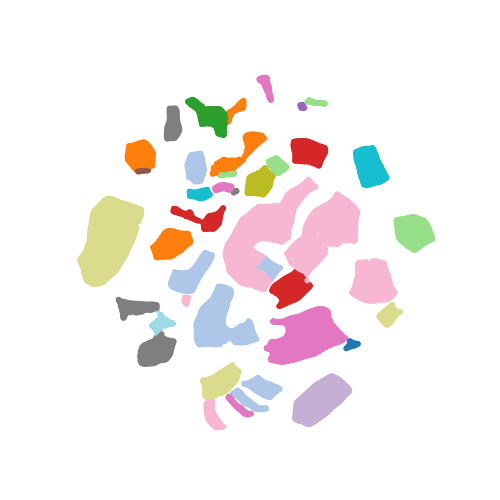}}
\end{minipage}
\begin{minipage}[t]{0.24\linewidth}
\centering
\subfloat[Cross-VAE view feature]{
\includegraphics[scale=0.4]{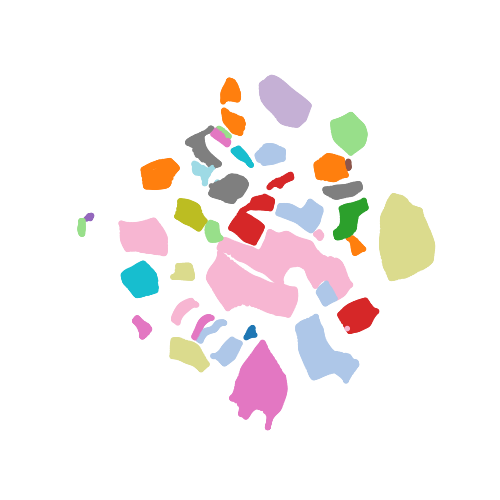}}
\end{minipage}
\begin{minipage}[t]{0.24\linewidth}
\centering
\subfloat[Standard AAE view feature]{
\includegraphics[scale=0.4]{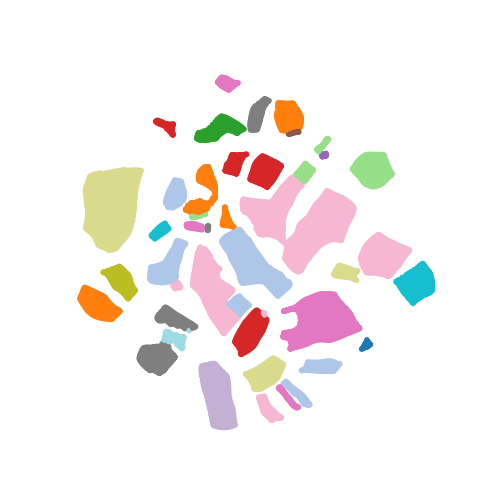}}
\end{minipage}
\begin{minipage}[t]{0.24\linewidth}
\centering
\subfloat[Cross-AAE view feature]{
\includegraphics[scale=0.4]{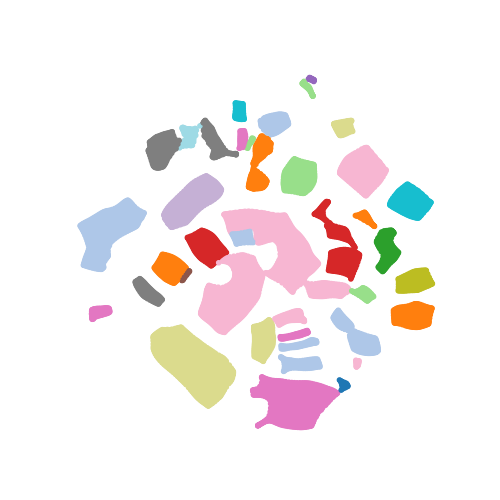}}
\end{minipage}
\caption{t-SNE visualization of features in the IP dataset.}
\label{fig_tsne_vae}
\end{figure*} 
\subsection{Cross representation ablation experiments}
One of the purposes of our cross-prediction design is to construct a pair of views for our contrastive learning process. Another purpose is to learn a contrastive feature implicitly. 

\begin{table}[tb]
\caption{Ablation experiments of contrastive learning}
\begin{tabular}{ccc|c}
\hline
\textbf{ } & \textbf{Cross-VAE}& \textbf{Cross-AAE}& \textbf{Contrastive}\\
\hline
AA($\%$)& 93.85& 95.71& \textbf{97.89}\\
OA($\%$)& 91.33& 92.13& \textbf{98.20}\\
\hline
\end{tabular}
\begin{tablenotes}
    \footnotesize
    \item{Data in bold are the best performance.}
\end{tablenotes}
\label{Cross_contrast_ab}
\end{table}

Here we do the ablation experiments for this cross representation learning method to validate the effects of our cross-prediction design. We expect it could bring extra benefits of better contrastive features. We directly apply an SVM over the contrastive features from our cross-prediction tasks and standard autoencoders. We use the same channel numbers ($15$ channels) to predict features in the standard representation learning for a fair comparison, which uses less information from the implementation in ContrastNet. We used the IP dataset (10$\%$ data used in the training process) to show classification performance.

Table \ref{Cross_ab} shows classification experiment results from cross-representation features and standard autoencoders features. It shows our contrastive features also have better performance than standard features. It means, when making features a transferable pair for contrastive learning tasks, the cross-representation style also improves the performance. For VAE, it has a significant improvement. For AAE, both methods work well.

We also visualize the features from the standard autoencoder and our cross-prediction method by t-SNE method. From Figure \ref{fig_tsne_vae}, we can see features from our method are more compact. For VAE, some classes of our cross-VAE features are more discriminative, which shows the advantage of our methods in the VAE pipeline. For AAE, both work well. Features from our methods are more compact and discriminative too.

\subsection{Contrastive learning ablation experiments}
With better features extracted from the cross-VAE view and the cross-AAE view (Tabel \ref{Cross_ab}), we apply a contrastive learning method over these two feature views to get better features in the IP dataset (10$\%$ data used in the training process). Results is shown in Table \ref{Cross_contrast_ab}.

From Table \ref{Cross_contrast_ab}, we can see we get better features with our contrastive methods. In some sense, our method fuses the features from cross-AAE and cross-VAE, which highly improve the performances of the HSI classification task. Higher performance shows our method can extract consistently semantic information for HSI classification downstram tasks.

\subsection{Detailed contrastive learning experiments of other datasets}
\begin{table*}[tbh]
\begin{center}
\begin{tabular*}{\textwidth}{@{\extracolsep{\fill}}ccccccccc}
\hline
\textbf{Class}&\multicolumn{3}{c}{\textbf{Supervised Feature Extraction}}&\multicolumn{5}{c}{\textbf{Unupervised Feature Extraction}} \\
\hline
\textbf{ } & \textbf{SLGDA}& \textbf{1D-CNN}& \textbf{S-CNN}& \textbf{3DCAE}& \textbf{AAE}& \textbf{VAE}& \textbf{ContrastNet}&\textbf{Ours model} \\
\hline
0 & 98.14 &97.98& 99.55& 100.00& 99.98& 99.74 &99.93&\textbf{100.00} \\
1 & 99.44 &99.25& 99.43& 99.29& 100.00& 99.79 &99.80&\textbf{100.00} \\
2 & 99.29 &94.43& 98.81& 97.13& 100.00& 100.00& 99.95&\textbf{100.00} \\
3 & 99.57& 99.42& 97.45& 97.91& 99.09& 99.57& 98.01&\textbf{100.00} \\
4 & 98.06 &96.60& 97.96& 98.26& 99.42& 99.71& 99.48&\textbf{99.92} \\
5 & 99.32 &99.51& 99.83& \textbf{99.98}& 99.97& 99.86& 99.94&99.89\\
6 & 99.33 &99.27& 99.59& 99.64& 99.93& 99.96& 99.79&\textbf{100.00}\\
7 &89.48 &86.79 &94.40& 91.58& 91.21 &86.91 &99.53&\textbf{99.89}\\
8 & 99.65 &99.08& 98.85& 99.28 &99.69& 99.99 &99.71&\textbf{100.00}\\
9 & 97.94 &93.71& 97.35& 96.65 &98.46& 96.54 &99.80&\textbf{99.81}\\
10 & 99.06 &94.55& 97.71&97.74 &99.57& 100.00& 99.80&\textbf{99.90}\\
11 & 100.00& 99.59& 98.73&98.84 &100.00& 99.44& 99.98&\textbf{100.00}\\
12 & 97.82& 97.50 &96.72&99.26& \textbf{99.89} &97.24& 98.20&96.32\\
13 & 90.47& 94.08 &95.22&97.49& 99.08 &96.49& 98.62&\textbf{100.00}\\
14 & 69.51& 66.52 &95.61&87.85& 93.69 &87.90& 99.53&\textbf{99.55}\\
15 & 99.00& 97.48 &99.44&98.34& 99.73 &99.61& 99.57&\textbf{100.00}\\
\hline
AA($\%$)& 96.01& 94.73& 97.39& 97.45 &98.73& 97.67 &99.48&\textbf{99.70}\\
OA($\%$)& 93.31& 91.30& 97.92&95.81& 97.10 &95.23& 99.60&\textbf{99.82}\\
\hline
\end{tabular*}
\caption{Detailed evaluation in Salinas dataset}
\label{tabSA_detail}
\end{center}
\end{table*}

\begin{table*}[tbh]
\begin{center}
\begin{tabular*}{\textwidth}{@{\extracolsep{\fill}}ccccccccc}
\hline
\textbf{Class}&\multicolumn{3}{c}{\textbf{Supervised Feature Extraction}}&\multicolumn{5}{c}{\textbf{Unupervised Feature Extraction}} \\
\hline
\textbf{ } & \textbf{SLGDA}& \textbf{1D-CNN}& \textbf{S-CNN}& \textbf{3DCAE}& \textbf{AAE}& \textbf{VAE}& \textbf{ContrastNet}&\textbf{Ours model} \\
\hline
0 & 94.66& 90.93& 95.40& 95.21& 95.39& 81.31& 99.49&\textbf{100.00}\\
1 & 97.83 &96.94 &97.31& 96.06 &98.96 &97.65 &99.98&\textbf{100.00} \\
2 & 77.27 &69.43 &81.21& 91.32 &97.49 &91.00 &99.06&\textbf{99.74} \\
3 & 93.18 &90.32 &95.83& 98.28 &96.94 &94.79 &97.75&\textbf{99.38} \\
4 & 98.51& 99.44& 99.91& 95.55& 99.94& 99.94& 99.81&\textbf{100.00} \\
5 & 90.08 &73.69& 95.29& 95.30& 99.29& 99.91& 99.90&\textbf{100.00}\\
6 & 85.34 &83.42 &87.05& 95.14&\textbf{100.00}& 99.42& 99.83&99.42\\
7 &90.49 &83.65 &87.35& 91.38 &97.90 &94.77 &98.79&\textbf{99.37}\\
8 &99.37& 98.23& 95.66& \textbf{99.96}& 96.32& 88.62& 94.84&95.77\\
\hline
AA($\%$)& 91.86& 87.34& 94.75 &95.36& 98.03& 94.16 &98.83&\textbf{99.30}\\
OA($\%$)& 94.15& 89.99& 92.78& 95.39& 98.14& 94.53& 99.46 &\textbf{99.78}\\
\hline
\end{tabular*}
\caption{Detailed evaluation in University of Pavia dataset}
\label{tabPU_detail}
\end{center}
\end{table*}

Table \ref{tabSA_detail} and Table \ref{tabPU_detail} shows details of our experiments in SA and PU datasets. Because each class's $OA$ and $AA$ performance are very close, we did not put the details of the experiments in the paper. So we list the detailed performance in the appendix for further analysis. Data in bold are the best performance.

Table \ref{tabSA_detail} shows experiments results in SA dataset.  We can see our model also gets state-of-the-art performance with the highest OA and highest AA. It achieves the best performance for overlapping data (such as classes $10, 11, 12, 13$, where other methods may not distinguish well). Our method gets the highest performance from contrastive features. It shows evidence that our cross-representation learning and contrastive strategy can extract better features. For the class $12$, performance from our method decreases a bit. It may be because it is hard to get contrastive features compared with similar classes $10,11,13$ (all lettuce-romaine).

Table \ref{tabPU_detail} shows experiment results in the PU dataset. We can see our model gets state-of-the-art performance with the highest OA and highest AA. And in most classes, it achieves the best performance. Our method gains the best performance and outperforms all other existing unsupervised methods. This impressive result shows features from our models have better representation ability. For the class $8$, performance from our method decreases a bit. It may come from `Shadows' data itself. For that, `Shadows' data has no distinguishable features. It is hard to distinguish `Shadows' compared with other semantic things, which may confuse contrastive feature extractors. Some dark items may disturb the learning process in our contrastive learning methods. Similarly, AAE and VAE also work well in this scenario. Our approach can further enhance the performance of standard auto-encoders and optimize the trade-off between AAE and VAE. 

\section{Conclusions}
In this paper, we proposed a cross-view-prediction style method for HSI feature learning. This style combines the cross-representation learning method and the contrastive learning method. We use this combination to fuse the advantage of the cross-representation learning method (detail texture learning) and contrastive learning method (high-level information learning).

For the cross-representation learning method, we design four typical cross-channel-prediction strategies as augmentation methods to construct different views of original data. Our cross representation learning method naturally suits handling HSI data with highly redundant channels. And this unsupervised cross-channel-prediction task implicitly explores contrastive features in hyperspectral space. For the contrastive learning method, we propose an interpretable and straightforward contrastive learning method to optimize the features further. More representative and compact features derive from our contrastive learning pipeline, which maximizes mutual information and minimizes conditional entropy across different views. Based on these two methods, our cross-view-prediction style tends to extract better features with consistent semantics for hyperspectral classification tasks. Our experiments show our proposed method outperforms all existing hyperspectral image classification algorithms.

\newpage

\bibliographystyle{unsrt}
\bibliography{main}
\end{document}